\documentclass{article} % For LaTeX2e
\usepackage{preprint}

\usepackage{microtype}
\usepackage{hyperref}
\usepackage{url}
\usepackage{booktabs}
\usepackage{tikz}

\usepackage{multirow}

\usepackage{amsmath,amsfonts,bm}

\def\eqref#1{equation~\ref{#1}}
\def\1{\bm{1}}

\DeclareMathAlphabet{\mathsfit}{\encodingdefault}{\sfdefault}{m}{sl}
\SetMathAlphabet{\mathsfit}{bold}{\encodingdefault}{\sfdefault}{bx}{n}

\def\gY{{\mathcal{Y}}}

\DeclareMathOperator*{\argmin}{arg\,min}

\usepackage{amssymb, comment}

\usepackage[ruled,vlined]{algorithm2e}

\usepackage{scalerel,xparse}

\title{Investigating Regularization of Self-Play Language Models }

\author{Reda Alami\thanks{Equal contribution}, Abdalgader Abubaker$^{*}$, Mastane Achab, Mohamed El Amine Seddik\\
\& \textbf{Salem Lahlou}  \\
Technology Innovation Institute,  9639 Masdar City, Abu Dhabi, United Arab Emirates \\\texttt{\{name.surname\}@tii.ae} 
}

\newtheorem{remark}{Remark}

\colmfinalcopy % Uncomment for camera-ready version, but NOT for submission.
\begin{document}

\maketitle

\begin{abstract}
This paper explores the effects of various forms of regularization in the context of language model alignment via self-play.
%We introduces a new variant of the self-play fine-tuning (SPIN) algorithm for the problem of large language model (LLM) alignment from pairwise preferences.
While both reinforcement learning from human feedback (RLHF) and direct preference optimization (DPO) require to collect costly human-annotated pairwise preferences, the self-play fine-tuning (SPIN) approach replaces the rejected answers by data generated from the previous iterate.
However, the SPIN method presents a performance instability issue in the learning phase, which can be mitigated by playing against a mixture of the two previous iterates.
%a risk of deviating too much from the initial reference policy.
In the same vein, we propose in this work to address this issue from two perspectives: first, by incorporating an additional Kullback-Leibler (KL) regularization to stay at the proximity of the reference policy; second, by using the idea of fictitious play which smoothens the opponent policy across all previous iterations.
In particular, we show that the KL-based regularizer boils down to replacing the previous policy by its geometric mixture with the base policy inside of the SPIN loss function.
% Nevertheless, the SPIN loss function uses a Kullback-Leibler (KL) regularization with respect to \emph{only the previous iterate}.
% The present work proposes to fix this data/regularization inconsistency by using a KL-penalty obtained via either an arithmetic or a geometric mixture of the previous model and the reference policy.
%
We finally discuss empirical results on MT-Bench as well as on the Hugging Face Open LLM Leaderboard.
\end{abstract}

\section{Introduction}

Large Language Models (LLMs) have shown remarkable abilities in a broad spectrum of fields that demand complex reasoning and in-depth expertise. These models are adept at navigating tasks like solving mathematical problems \citep{GSM8k}, generating code \citep{li2022competition}, producing text \citep{touvron2023llama}, summarizing documents, and crafting creative works, to name a few.

A notable advancement in the development of LLMs is their post-pretraining refinement to encourage more favorable behaviors, a process that often depends on expensive human-curated data. Common strategies for this refinement include Supervised Fine-Tuning (SFT) \citep{ouyang2022training,tunstall2023zephyr}, which leverages human annotated prompt-response examples, and on the other hand Reinforcement Learning from Human Feedback (RLHF) \citep{christiano2017deep,ziegler2019fine,bai2022training}, Direct Preference Optimization (DPO) \citep{rafailov2023direct}, Identity Preference Optimization \citep[IPO;][]{IPO}, Sequence Likelihood Calibration with Human Feedback \citep[SLIC;][]{zhao2023slic}, which utilize human preferences.

Recently, \cite{chen2024self} introduced a pioneering fine-tuning approach named Self-Play fIne-tuNing (SPIN), starting with a base model. SPIN engages the LLM in a self-play format, thereby removing the need for expert annotation from either humans or superior LLMs such as GPT-4 \citep{achiam2023gpt}. Specifically, using the LLM at a previous state $t$, denoted as $\pi_{\theta_t}$, this method generates responses to prompts $x$ obtained from a human-annotated Supervised Fine-Tuning (SFT) dataset, and uses the generated text to train a new LLM, $\pi_{\theta_{t+1}}$, to identify the responses made by $\pi_{\theta_t}$ as distinct from those made by humans. This setup resembles a two-player game where the main player, the new LLM $\pi_{\theta_{t+1}}$, aims to differentiate between opponent $\pi_{\theta_t}$ 's responses and those created by humans. The updated LLM, $\pi_{{\theta}_{t+1}}$, is refined from $\pi_{{\theta}_t}$ to favor responses closer to the actual data distribution $\pi_{\text {data }}$ over those from $\pi_{{\theta}_t}$, leading to a model, $\pi_{{\theta}_{t+1}}$, that better matches $\pi_{\text {data }}$. In successive iterations, the newly refined LLM $\pi_{\theta_{t+1}}$ then serves as the opponent in generating responses, where the goal is to drive the LLM towards convergence with $\pi_{{\theta}^*}=\pi_{\text {data }}$ (assuming such parameter ${\theta}^*$ exists), achieving a state where the most advanced LLM cannot distinguish its previous version's responses from those made by humans.

Nonetheless, the official SPIN implementation differs from the proposed algorithm as it relies on a mixture of the previous two iterates as a generator\footnote{\url{https://github.com/uclaml/SPIN/tree/main}}. Allegedly, this introduces some form of stability, as it ensures that the model performance doesn't significantly deviate from its previous iterates.

\begin{figure}[t!]
    \centering
\input{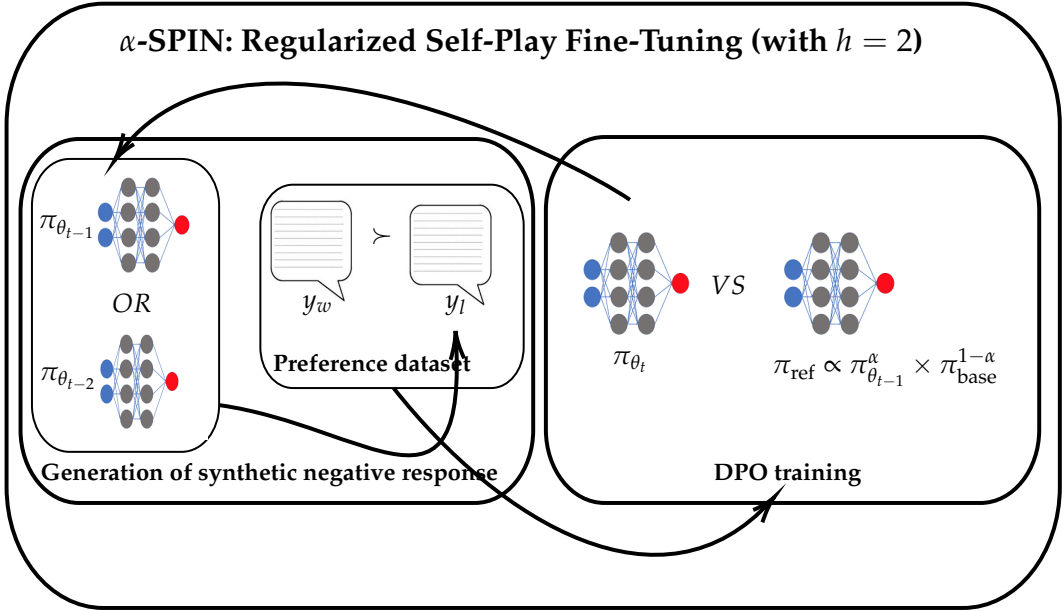}
    \caption{Framework of regularized self-play fine tuning (Algorithm \ref{alg:alphaSPIN}). The pair context/winner-answer is picked from the SFT dataset: $(x,y_w) \in \mathcal{D}_{\text{SFT}}$. The negative response $y_l$ is generated according to either the previous policy $\pi_{\theta_{t-1}}$ or a mixture of the previous policies. 
    The fine-tuning of the model $\pi_{\theta_{t}}$ is done using the maximum likelihood estimation in the DPO approach against the reference model $\pi _{\text{ref}}
\propto  \pi _{\theta _{t-1}}^{\alpha} \times  \pi _{\text{base}}^{1-\alpha}$ for a given $\alpha 
\in (0,1)$. 
}
    \label{fig:proposed-approach}
\end{figure}

% Nonetheless, SPIN suffers from an inherent instability, that makes the model deviate from partially mitigated by using a mixture of the previous two iterates as a generator.
In this paper, we investigate several regularization techniques for mitigating this instability issue of SPIN. 
We propose (1) an adaptation of the SPIN framework by introducing in the loss function an additional KL-regularizer to stay in the proximity of the base model, and (2) alternative sampling schemes that differ in how they account for the previous iterates of the LLM. 
%The original SPIN implementation gradually diverges from the SFT model as it evolves, potentially losing some of the valuable behaviors that have been initially captured. 
% Our enhanced approach seeks to maintain a closer alignment with the base model by incorporating it directly into the iterative process. By doing so, we aim to create a more effective learning procedure where the new LLM not only learns to differentiate between its past self and human responses but also retains a strong connection to the base model. This adaptation is designed to mitigate the risk of deviation from the desirable characteristics embedded in the base model, thus getting a more controlled and guided fine tuning of the LLM. Our approach is depicted in Figure \ref{fig:proposed-approach}. 
Our proposed algorithm, $\alpha-$SPIN, depicted in Figure \ref{fig:proposed-approach}, allows us to investigate the extent to which maintaining a closer alignment with the base model, by incorporating it into the iterative process, positively affects the learning procedure.  This adaptation is designed to mitigate the risk of deviation from the desirable characteristics embedded in the base model, thus getting a more controlled and guided fine-tuning of the LLM. Additionally, $\alpha-$SPIN allows us to evaluate the effect of the sampler of the ``loser'' responses, as it involves a history length parameter $h$, controlling for the number of past iterates used to create the averaged opponent. Additionally, we investigate in appendix \ref{sec::gflownets} an alignment of the generator with the reference model used in the loss function, using GFlowNet-finetuning \citep{hu2023amortizing}, in order to obtain a sampler from a geometric mixture.

\paragraph{Outline}
The remainder of the paper is organized as follows. Section \ref{sec::relatedWork} describes related work on aligning large language models with human preference. Then, section \ref{sec::proposedMethod} introduces our $\alpha$-SPIN framework that allows us to compare different regularization schemes. In section \ref{sec::evaluation}, we conduct evaluations over MT-Bench and Hugging Face Open LLM Leaderboard.
%Finally, section \ref{sec::conclusion} concludes the paper with a list of future work.

\paragraph{Notations}
We denote by $\sigma: \mathbb{R} \rightarrow (0,1)$ the sigmoid function given by $\sigma(z)=(1+e^{-z})^{-1}$.
For two distributions $\mu, \nu$ in the probability simplex $\Delta_K$ ($K\ge 1$), with $\mu$ absolutely continuous with respect to $\nu$, the Kullback-Leibler divergence is equal to $\text{KL}(\mu \| \nu) = \sum_{k=1}^K \mu_k \log \frac{\mu_k}{\nu_k}$.
Given a ``state'' space $\mathcal{X}$ and an ``action'' space $\mathcal{Y}$ (both assumed to be finite for simplicity), we will call ``policy'' any mapping $\pi : \mathcal{X} \rightarrow \Delta_{|\mathcal{Y}|}$, i.e. for any state $x$, $\pi(\cdot | x)$ is a distribution over $\mathcal{Y}$. Throughout this paper, a state $x$ will correspond to a prompt, with $y \sim \pi(\cdot | x)$ an answer produced by some language model represented as a policy $\pi$.
Given two vectors $p,q \in \mathbb{R}^n$, we denote by $p \odot q \in \mathbb{R}^n$ their entrywise Hadamard product ; if $p \in \mathbb{R}^n_+$ and $\alpha \ge 0$, then $p^\alpha$ denotes the vector obtained by elevating each entry of $p$ to the power $\alpha$.

\section{Related work}
\label{sec::relatedWork}
Let us first recall the RLHF paradigm, which assumes that we already have a pre-trained base model $\pi_{\text{base}}$, and pairwise comparisons data, i.e. a set of triplets $(x, y_w, y_l)$ with $y_w,y_l \sim \pi_{\text{ref}}(\cdot | x)$ two answers to the same prompt $x$, such that ``$y_w \succ y_l$'' meaning that $y_w$ was preferred over $y_l$ by some human annotator.

% \textcolor{red}{ADD REFERENCES}
% \begin{itemize}
%     \item \cite{casper2023open} \textcolor{red}{[Open problems and fundamental limitations of reinforcement learning from human feedback]}
%     \item \cite{tiapkin2023regularized} \textcolor{red}{[Demonstration-Regularized RL]}
% \end{itemize}

\begin{comment}

\paragraph{Identity preference Optimization (IPO) }

Introduced by \cite{IPO}, ... 

\begin{equation}
\mathcal{L}_{\text{IPO}}(\theta) =
\mathbb{E}_{x, y_w \succ y_l}\left[ \log \left(\frac{\pi_\theta(y_w \mid x) \pi_{\mathrm{ref}}\left(y_l \mid x\right)}{\pi_\theta\left(y_l \mid x\right) \pi_{\mathrm{ref}}(y_w \mid x)}\right)-\frac{\tau^{-1}}{2}  \right]^2
\end{equation}

\paragraph{Sequence Likelihood Calibration with
Human Feedback (SLiC) }

Introduced by \cite{zhao2023slic}, ... 

\begin{equation}
\mathcal{L}_{\text{SLiC}}(\theta) = \mathbb{E}_{x, y_w \succ y_l}\left[ \max \left(0, \delta-\log \pi_\theta\left(y_w \mid \mathbf{x}\right)+\log \pi_\theta\left(y_l \mid \mathbf{x}\right)\right)-\lambda \log \pi_\theta\left(\mathbf{y}_{\mathrm{ref}} \mid \mathbf{x}\right) \right]
\end{equation}

\paragraph{Other possible variants}

\textcolor{red}{[See General Preference  Optimization paper from DeepMind]} 
\end{comment}

\subsection{Reinforcement Learning from Human Feedback (RLHF)}

RLHF relies on the classic Bradley-Terry (BT) model \citep{bradley1952rank} for pairwise ranking probabilities. More precisely, it assumes the existence of a reward function $r^*$ such that, given any pair of answers $y,y'$ to the same prompt $x$, the pairwise probability of $y$ being preferred over $y'$ satisfies:
\begin{equation}
    \tag{BT}
    \label{eq:BT}
    p^*( y \succ y' | x) = \sigma( r^*(x,y) - r^*(x, y') ) .
\end{equation}
Given this BT assumption, the first phase of RLHF consists in approximating the (unknown) reward function $r^*$ by a parametric reward model $r_\phi$ that is learned via maximum likelihood, i.e. by solving:
\begin{equation}
\label{eq:reward_learning}
    \min_{\phi} - \mathbb{E}_{x,y_w \succ y_l}\left[ \log \sigma(r_\phi(x,y_w) - r_\phi(x,y_l)) \right] .
\end{equation}

Then, the second phase of RLHF aims at learning a parametric policy $\pi_\theta$ that maximizes the reward model $r_{\hat \phi}$ (from the first phase), while still staying in the proximity of the reference model through a KL-penalty:
\begin{equation}
    \max_\theta \mathbb{E}_x \left[ \mathbb{E}_{y \sim \pi_\theta(\cdot|x)}[ r_{\hat \phi}(x,y) ] - \beta \text{KL}( \pi_\theta(\cdot | x) \| \pi_{\text{ref}}(\cdot | x) ) \right] .
\end{equation}
The most popular method for solving this problem is the PPO algorithm \citep{schulman2017proximal} applied with the reward function $\tilde r(x,y) = r_{\hat \phi}(x,y) - \beta \log \frac{\pi_\theta(y|x)}{\pi_{\text{ref}}(y|x)} $.

\subsection{Direct Preference Optimization (DPO)}

In \cite{rafailov2023direct}, the authors proposed to bypass the reward learning phase from RLHF (Eq.\ref{eq:reward_learning}) by leveraging the fact that, for any given reward function $r$, 
the KL-regularized problem
\begin{equation}
\label{eq:kl_reg}
    \max_{\pi} \mathbb{E}_{x}\left[ \pi(\cdot|x)^\intercal r(x,\cdot ) - \beta \text{KL}(\pi(\cdot|x) \| \pi_{\text{ref}}(\cdot|x)) \right]
\end{equation}
admits the following closed-form solution:
\begin{equation}
\label{eq:pi_r}
    \pi_r(y | x) = \frac{1}{Z(x)} \pi_{\text{ref}}(y | x) \exp( \frac{1}{\beta} r(x,y) ) ,
\end{equation}
where the partition function $Z(x)=\sum_y \pi_{\text{ref}}(y | x) \exp( \frac{1}{\beta} r(x,y) )$ is in general challenging to compute.
In other words, there exists an explicit mapping between $r$ and the corresponding optimal policy $\pi_r$. Indeed, by taking the logarithm on both sides of Eq.(\ref{eq:pi_r}), we obtain after rearranging the terms:
\begin{equation}
    \label{eq:r_pi}
    r(x,y) = \beta \log \frac{\pi_r(y|x)}{\pi_{\text{ref}}(y|x)} + \beta \log Z(x) .
\end{equation}
DPO implicitly uses the reward from Eq.(\ref{eq:r_pi}) and hence does not need to learn it as in RLHF.
Luckily, by re-injecting this expression into the Bradley-Terry likelihood, the term involving $Z(x)$ cancels in the difference of the rewards, and we obtain the DPO loss:
\begin{equation}
\label{eq:dpo_loss}
    \mathcal{L}_{\text{DPO}}(\theta ; \pi_{\text{ref}}) = - \mathbb{E}_{x, y_w \succ y_l}\left[ \log \sigma\left( \beta \log\frac{\pi_{\theta}(y_w|x)}{\pi_{\text{ref}}(y_w|x)} - \beta \log\frac{\pi_{\theta}(y_l|x)}{\pi_{\text{ref}}(y_l|x)} \right) \right] .
\end{equation}

\subsection{Alignment methods based on alternative loss functions}
\label{subsec:alternative}

Let us recall alternative methodologies and loss functions used in the alignment of large language models, highlighting two notable techniques developed in recent research.

\noindent \emph{Identity Preference Optimization (IPO).}
\cite{IPO} introduced IPO, a novel approach aimed at enhancing model alignment through preference-based optimization. 
The idea behind IPO is similar to DPO, except that it replaces the logistic regression loss by a least squares loss.
%The underlying principle of IPO is to adjust the model's output distribution to closely align with a reference distribution, particularly when distinguishing between high and low preference outputs.
Formally, given a parameter $\tau > 0$, the loss function of IPO, is defined as follows:  
\begin{equation*}
\mathcal{L}_{\text{IPO}}(\theta ; \pi_{\text{ref}}) = \mathbb{E}_{x, y_w \succ y_l} \left[  \left(\log \frac{\pi_\theta(y_w | x) }{ \pi_{\text{ref}}(y_w | x)} - \log \frac{\pi_\theta(y_l | x) }{ \pi_{\text{ref}}(y_l | x)} -
\frac{\tau^{-1}}{2}\right)^2 \right] .
\end{equation*}
%This formulation aims to minimize the squared difference between the logarithmic ratio of the probabilities assigned by the model $\left(\pi_\theta\right)$ and the reference model $\left(\pi_{\text {ref }}\right)$ to the preferred $\left(y_w\right)$ and less preferred $\left(y_l\right)$ outputs, adjusted by a term inversely proportional to $\tau$.
In particular, IPO has been shown to be less prone to overfitting than RLHF and DPO.

\noindent \emph{Sequence Likelihood Calibration with Human Feedback (SLiC).}
\cite{zhao2023slic} developed the SLiC technique, which calibrates the likelihood of sequence outputs using direct human feedback. The SLiC methodology introduces a structured approach to modifying the model's behavior by penalizing deviations from expected outcomes, as indicated by human preferences. The loss function for SLiC, incorporating parameters $\delta>0$ and $\lambda>0$, is expressed as:
\begin{equation*}
\mathcal{L}_{\text{SLiC}}(\theta) = \mathop{\mathbb{E}}_{(x,y_{\text{ref}})\in \mathcal{D}_{\text{SFT}}, y_w \succ y_l  }  \left[ \max \left(0, \delta-\log \pi_\theta\left(y_w | x\right)+\log \pi_\theta\left(y_l | x \right)\right)-\lambda \log \pi_\theta\left(y_{\text{ref}} | x \right) \right] .
\end{equation*}
Here, the function seeks to adjust the model's probability distribution $\left(\pi_\theta\right)$ such that it aligns with the preferences indicated by human feedback, while also maintaining a balance between rewarding desirable outcomes and penalizing deviations from these expectations.

\subsection{Self-Play Fine-Tuning (SPIN)}

An important requirement of RLHF and DPO is the human-annotated comparison data, which can be very costly to acquire.
The SPIN algorithm \citep{chen2024self} is an iterative method that only requires an SFT dataset $\mathcal{D}_{\text{SFT}}$ composed of prompt-answer pairs $(x,y)$.
Indeed, at each iteration $t$, the next SPIN iterate $\pi_{\theta_t}$ is obtained by minimizing the DPO loss with the winner answer $y_w$ picked from some \emph{real} SFT data, while the loser answer $y_l$ is \emph{generated} from the previous policy $\pi_{\theta_{t-1}}$.

\begin{algorithm}[H]
\SetAlgoLined
%\KwResult{Find the greatest common divisor of two numbers}
\KwIn{SFT data $\mathcal{D}_{\text{SFT}}$, base policy $\pi_{\text{base}}$\,.}
Set $\forall k < 0$ : $\pi_{\theta_{k}} = \pi_{\text{base}}$

\For{$t = 0$ \KwTo $T$}{
  Set \textcolor{blue}{$\pi_{\text{ref}} = \pi_{\theta_{t-1}}$}
  
  Gather triplets $(x,y_w,y_l)$ with $(x,y_w)\in \mathcal{D}_{\text{SFT}}$ and \textcolor{blue}{$y_l \sim  \pi_{\theta_{t-1}}(\cdot|x) $ } \\ % \textcolor{blue}{$y_l \sim \frac12 ( \pi_{\theta_{t-1}}(\cdot|x) + \pi_{\text{base}}(\cdot|x) )$ } \; %\textcolor{blue}{$y_l \sim \pi_{\theta_{t-1}}(\cdot|x)$ or $y_l \sim \pi_{\text{base}}(\cdot|x)$}\;
  Minimize DPO loss (Eq.\ref{eq:dpo_loss}): $\theta_t = \argmin_\theta \mathcal{L}_{\text{DPO}}(\theta ; \pi_{\text{ref}})$
}
\KwOut{Final SPIN policy $\pi_{\theta_T}$}
\caption{SPIN \citep{chen2024self} }
\label{alg:spin}
\end{algorithm}

\begin{remark}[Official implementation of SPIN]
\label{rmk:official_implem}
    We point out that the SPIN Algorithm \ref{alg:spin} (as formally described in \cite{chen2024self}) differs from its official implementation. Indeed, at each iteration, the latter evenly mixes the two previous policies for the generation of the negative answers, i.e. \textcolor{blue}{$y_l \sim  \frac12 \pi_{\theta_{t-1}}(\cdot|x) + \frac12 \pi_{\theta_{t-2}}(\cdot|x) $ }.
\end{remark}

Motivated by the two-step averaging trick discussed in Remark \ref{rmk:official_implem}, that the authors interpret as a form of regularization\footnote{See \url{https://github.com/uclaml/SPIN/issues/11}}, we propose in the next section a general framework for regularizing the SPIN method and further investigate different regularization strategies.

%\section{Proposed method: $\alpha$-SPIN / Geometric Self-Play (GSP) / Geometric Self-Play Alignment (G-SPA)}

%\section{Problem Formulation}
\section{Regularization of self-play fine-tuning}
\label{sec::proposedMethod}

In our investigation of regularization of self-play fine-tuning, we propose two complementary directions to generalize the SPIN algorithm. We first start by describing these two directions, summarizing them in the  $\alpha-$SPIN algorithm in \ref{alg:alphaSPIN}. Our experimental investigation in Section \ref{sec::evaluation} considers different settings of $\alpha-$SPIN.

%We first start by describing the two modifications brought to SPIN, which lead to the $\alpha-$SPIN algorithm in \ref{alg:alphaSPIN}.

% We now introduce our new SPIN-based method. 
%It differs from the original SPIN in the unified way in which we choose the reference policy to be the \emph{same} as the policy that generates the rejected answers.
\begin{comment}
We suggest three options for choosing the reference model at each iteration:
\begin{enumerate}%[(1)]
    \item either as an \emph{arithmetic mixture} between $\pi_{\theta_{t-1}}$ and $\pi_{\text{base}}$ ;
    \begin{itemize}
        \item \textcolor{red}{Mixing the probabilities ... }
    \end{itemize}
    \item or, proportional to a \emph{geometric mixture} between $\pi_{\theta_{t-1}}$ and $\pi_{\text{base}}$ ;
        \begin{itemize}
        \item \textcolor{red}{Mixing the probabilities ... }
    \end{itemize}
    \item or, by computing an \emph{exponential moving average (EMA)} in the parameter-space (i.e. on the $\theta$'s).
        \begin{itemize}
        \item \textcolor{red}{Mixing the parameters of both the model (Creation of an intermediate LLM)}
    \end{itemize}
\end{enumerate}
\end{comment}

\subsection{From KL regularization to geometric mixture}

At a given iteration of SPIN, we optimize the DPO loss in Eq.(\ref{eq:dpo_loss}), which we recall is derived from Eq.(\ref{eq:kl_reg}) with $\pi_{\text{ref}} = \pi_{\theta_{t-1}}$. To stay in the proximity of the base model, we propose to add an additional KL regularizer with respect to the base model. This amounts to replacing $\text{KL}(\pi(\cdot|x) \| \pi_{\theta_{t-1}}(\cdot|x))$ in the loss of Eq.(\ref{eq:kl_reg}) with $\alpha \text{KL}(\pi(\cdot|x) \| \pi_{\theta_{t-1}}(\cdot|x)) + (1-\alpha) \text{KL}(\pi(\cdot|x) \| \pi_{\text{base}}(\cdot|x))$, for some $\alpha \in (0, 1)$. This corresponds to using a reference model which is a geometric mixture of the previous iterate and the base model: $\pi_{\text{ref}}(\cdot|x) \propto \pi_{\theta_{t-1}}(\cdot|x)^\alpha \odot \pi_{\text{base}}(\cdot|x)^{1-\alpha}$, given that 
\begin{equation*}
    \text{KL}(\pi(\cdot|x) \| \pi_{\text{ref}}(\cdot|x))
    = \alpha \text{KL}(\pi(\cdot|x) \| \pi_{\theta_{t-1}}(\cdot|x)) + (1-\alpha) \text{KL}(\pi(\cdot|x) \| \pi_{\text{base}}(\cdot|x)) + c(x),
\end{equation*}
where $c(x)$ is simply a normalization term independent of $\pi$. Notably, this geometric mixture component is also used in the Nash-MD method proposed by \cite{munos2023nash}.

\subsection{Fictitious play data generation}

Following the regularization initially introduced in the SPIN implementation (Remark \ref{rmk:official_implem}), which requires using an arithmetic mixture of the previous two iterates to generate synthetic data, we propose to investigate a generalization of this type of regularization by introducing a history length parameter $h$ in our proposed algorithm, that can take values larger than $2$. We also consider the special case of $h = \infty$, which corresponds to the fictitious play paradigm \citep{brown1951iterative}, for the synthetic data generation of our method.
For $h = \infty$, $\alpha$-SPIN performs a self-play against a uniform average over the history of the previous policies; at iteration $t\ge 1$, the negative answers are generated as follows:
$ y_l \sim \frac{1}{t} \sum_{0\le \tau \le t-1} \pi(\cdot|x) $.
%In other words, fictitious play in the context of synthetic data generation involves an iterative process where the $\alpha$-SPIN algorithm updates its strategy based on an aggregate of all past behaviors or strategies it has encountered or employed. 
This approach simulates a learning environment where the model essentially competes against a ``composite opponent'' that embodies the average of its historical strategies. 
This mechanism allows $\alpha$-SPIN to adapt and refine its strategy over time by considering a broad spectrum of past actions, rather than reacting to the most recent or a singular past strategy. 
It can be interpreted as a regularization/smoothing acting on the opponent's strategy.
In particular, this regularization interpretation has been developed in the literature (see e.g. \cite{cesa2003potential,shalev2012online,baudin2022smooth}) where a smooth version of the fictitious play is shown to be equivalent to an instance of the follow-the-regularized-leader \citep[FTRL;][]{mcmahan2011follow} algorithm.
%This process mimics human learning behaviors where strategies are often developed by reflecting on a wide range of past experiences and outcomes. %By synthesizing data through such a dynamic self-play mechanism, the model is encouraged to explore and optimize its strategy in a more comprehensive and nuanced manner, leading to improved alignment and performance.

\subsection{$\alpha$-SPIN algorithm}
We summarize our changes in the following algorithm:

\begin{algorithm}[H]
\SetAlgoLined
%\KwResult{Find the greatest common divisor of two numbers}
\KwIn{SFT data $\mathcal{D}_{\text{SFT}}$, base policy $\pi_{\text{base}}$, $\alpha \in (0,1)$, history length $h \geq 1$ .}
Set $\forall k < 0$ : $\pi_{\theta_{k}} = \pi_{\text{base}}$

\For{$t = 0$ \KwTo $T$}{
    Set \textcolor{blue}{$\pi_{\text{ref}}(\cdot|x) \propto \pi_{\theta_{t-1}}(\cdot|x)^\alpha \odot \pi_{\text{base}}(\cdot|x)^{1-\alpha} $} \tcp{Geometric Mixture}
    % \textcolor{blue}{$\pi_{\text{ref}}(\cdot|x) \propto \pi_{\theta_{t-1}}(\cdot|x)^\alpha \odot \pi_{\text{base}}(\cdot|x)^{1-\alpha} $}\;
    
  Gather triplets $(x,y_w,y_l)$ with $(x,y_w)\in \mathcal{D}_{\text{SFT}}$ and: 
  
\If{$h < \infty$}{
    \textcolor{blue}{$y_l \sim 
    \frac{1}{h} \sum_{\tau=t-h}^{ t-1} \pi_{\theta_\tau}(\cdot|x)
    $}
}
\ElseIf{$h = \infty$}{
    \textcolor{blue}{$y_l \sim \frac{1}{t} \sum_{0\le \tau \le t-1} \pi_{\theta_\tau}(\cdot|x) $} if $t \geq 1$, else \textcolor{blue}{$\pi_{\text{base}}(\cdot|x)$}
    \tcp{Fictitious Play}
}
  % \textcolor{blue}{$y_l \sim \alpha \pi_{\theta_{t-1}}(\cdot|x) + (1-\alpha) \pi_{\text{base}}(\cdot|x)$} %\textcolor{blue}{$y_l \sim \alpha \pi_{\theta_{t-1}}(\cdot|x) + (1-\alpha) \pi_{\text{base}}(\cdot|x)$ } 
  Minimize DPO loss (Eq.\ref{eq:dpo_loss}): $\theta_t = \argmin_\theta \mathcal{L}_{\text{DPO}}(\theta ; \pi_{\text{ref}})$
}
\KwOut{Final policy $\pi_{\theta_T}$}
%\caption{Geometric Self-Play Alignment (GSP / G-SPA)}
\caption{$\alpha$-SPIN}
\label{alg:alphaSPIN}
\end{algorithm}

Notice that in the limit case $\alpha=1$, $\alpha$-SPIN boils down to the original SPIN.
The other limit case $\alpha=0$ is similar to DPO with winning responses chosen in the SFT dataset instead of being generated.
For general $\alpha\in (0,1)$, $\alpha$-SPIN can be seen as an interpolation between these two limit cases by improving upon the previous iterate while still staying in the vicinity of the base policy.

\section{Performance evaluation}
\label{sec::evaluation}
This section provides an empirical analysis of $\alpha-$SPIN, which notably compares it to the original SPIN method. A specific care was given to the design of the experiments, in order to isolate the additional components and study their effects independently. 

In summary, our experimental results, detailed below, first  \textbf{confirm the positive effects brought by the KL regularization term}, at least on certain tasks. This was done by comparing $\alpha$-SPIN with $h=2$, to the official implementation of SPIN (that similarly uses $h=2$). Second, we \textbf{confirm the need for using a mixture in the sampler}, similar to SPIN (albeit not explicitly stated). This was done by comparing $\alpha$-SPIN with history length $h=2$ to $\alpha$-SPIN with history length $h=1$. Finally, we show that the \textbf{fictitious play approach shows promising results}, both with the KL regularization term in the loss ($\alpha$-SPIN) and without it (SPIN), and should be considered as a viable alternative to SPIN and its variants starting from the third iteration.

We point out that the presence of the reference policy $\pi_{\text{ref}}$ in the DPO loss (Eq.\ref{eq:dpo_loss}) originates in the KL regularization term $\text{KL}(\pi(\cdot|x) \| \pi_{\text{ref}}(\cdot|x))$ in Eq.(\ref{eq:kl_reg}). This is \emph{consistent} with the fact that the DPO approach generates the answers $y_w,y_l$ from $\pi_{\text{ref}}$. From this perspective, the SPIN iterative method appears to be \emph{inconsistent} since the standard implementation generates the rejected answers $y_l$ from an arithmetic mixture $\frac12 ( \pi_{\theta_{t-1}}(\cdot|x) + \pi_{\theta_{t-2}}(\cdot|x) )$ between the two previous iterates (see Remark \ref{rmk:official_implem}), but minimizes a DPO loss function that only takes the previous policy $\pi_{\text{ref}} = \pi_{\theta_{t-1}}$ as reference. Consequently, in order to design a ``consistent'' version of $\alpha$-SPIN, we would need to sample the negative answers from $\pi_{\text{ref}}(\cdot|x) \propto \pi_{\theta_{t-1}}(\cdot|x)^\alpha \odot \pi_{\text{base}}(\cdot|x)^{1-\alpha}$,
i.e., a geometric mixture between the previous iterate and the base policy. Therefore, in addition to the experiments presented in this section, we investigate in Appendix \ref{sec::gflownets} whether aligning the sampler of the loser answers $y_l$ with the reference model used in the loss function of $\alpha$-SPIN (the geometric mixture model), yields noticeable performance improvements. This poses an extra challenge, as explained in Appendix \ref{sec::gflownets}, that we tackle with GFlowNet-finetuning \citep{hu2023amortizing}. Our results show that more effort should be put into the specifics of the GFlowNet-finetuning.

\subsection{Experimental setup}
\textbf{Generation:} In this study, we use the \texttt{gemma-2b} model \citep{team2024gemma} as the base model, and UltraChat200k \citep{Ultrachat} as the SFT dataset, from which input prompts $x$ and winner answers $y_w$ are selected. Similar to SPIN, from the dataset of $200k$ examples, we sample $100k$ pairs $(x, y_w)$ at each iteration $t\ge 1$, while the corresponding $y_l$'s are sampled from each distinct model used to create the mixture model in Algorithm \ref{alg:alphaSPIN}, with equal probability. At the iteration $0$, we use $50k$ examples, where the loser answers are sampled from the base model. For example, in the fictitious play configuration, at iteration 3, we end up with $100k$ triplets $(x, y_w, y_l)$, where every previous iteration contributes with $33\%$. 

We note that the results reported in this section on SPIN differ from those of \cite{chen2024self}, as they consider a larger model (\texttt{zephyr-7b-sft-full}) for the base model. Given the limitations on the computational resources, and the purpose of this work being an investigation of regularization, we settled for the smaller \texttt{gemma-2b} base model in our experiments. The takeaways are however general and should apply to even larger models.

\textbf{Fine-tuning:} The fine-tuning is done according to the maximum likelihood process described in the DPO approach. All our experiments have been performed using the LoRA adapter \citep{hu2021lora} with rank $16$. Moreover, we apply the adapter into the \texttt{gemma-2b} model's projection layers with a dropout probability equal to $0.05$. All models were fine-tuned for $4$ epochs.

\textbf{Evaluation:} Similar to SPIN, we investigate the performances of the different configurations of $\alpha$-SPIN on the \textbf{MT-Bench} set of tasks. MT-Bench \citep{MTBench} is a benchmark designed by LMSYS to test the conversation and instruction-following capabilities of LLMs. It evaluates them through multi-turn conversations, focusing on their ability to engage in coherent, informative, and engaging exchanges. Additionally, we consider the \textbf{HuggingFace Open LLM Leaderboard} \citep{beeching2023open}, which comprises several challenging and diverse benchmarks, also designed to test various aspects of language understanding and reasoning abilities: AI2 Reasoning Challenge \citep[ARC;][]{ARC} focuses on testing the model's ability to reason through complex, science-based questions. "HellaSwag" \citep{Hellaswag} is designed to evaluate the model's understanding of context and its ability to predict the continuation of scenarios in texts and videos. "Winogrande" \citep{Winogrande} aims to assess the model's ability to resolve ambiguous pronouns in text, a task that requires both linguistic understanding and common-sense reasoning. "TruthfulQA" \citep{Truthfulqa} challenges models on their ability to provide truthful and factual answers, pushing the boundaries of veracity and knowledge in AI. "GSM8k" \citep{GSM8k} (Grade School Math 8k) tests the model's arithmetic and mathematical reasoning skills through a variety of grade-school level math problems. Lastly, "MMLU" (Massive Multitask Language Understanding \citep{MMLU}) is a comprehensive evaluation covering a broad range of subjects and disciplines, aiming to measure the model's general understanding across a wide array of topics.

We considered different values of $\alpha$ for the experiments and observed that for $\alpha<0.8$, the learned iterates stay too close to the base model, and their performances on the benchmarks do not significantly increase at each iteration. Therefore, in our experiments, we report the results for $\alpha=0.95$ only.

\subsection{Investigating the effect of KL regularization}
\begin{table}[t]
    \centering
    \caption{Performances of $\alpha$-SPIN ($h=1,2$) on the HuggingFace OpenLLM Leaderboard, for 3 iterations ($T=2$).}
    \vspace{0.5ex}
    \label{tab:h12}
    \resizebox{1\linewidth}{!}{
    \begin{tabular}{llccccccc}
        \toprule
        \multicolumn{2}{c}{} & ARC & TRUTHFULQA & WINOGRANDE & GSM8K & HELLASWAG & MMLU & \textbf{AVERAGE} \\
        \midrule
        %
        %\multirow{3}{*}{Iteration 0} & SPIN & $49.23$ & $32.97$ & $65.90$ & $18.73$ & $72.22$ & $41.06$ & $46.69$  \\
        \multirow{2}{*}{Iteration 0} & $\alpha-$SPIN ($h=2$) & $48.98$ & $33.00$ & $65.35$ & $19.33$ & $72.20$ & $41.01$ & $46.65$ \\
        & $\alpha-$SPIN ($h=1$) & $49.32$ & $33.15$ & $65.98$ & $18.20$ & $72.10$ & $41.03$ & $46.63$ \\
        \midrule
        %\multirow{3}{*}{Iteration 1} & SPIN & $49.32$ & $33.19$ & $66.46$ & $21.00$ & $73.06$ & $40.89$ & $47.32$  \\
        \multirow{2}{*}{Iteration 1} & $\alpha-$SPIN ($h=2$) & $49.32$ & $33.20$ & $66.22$ & $20.39$ & $73.06$ & $40.93$ & $47.19$ \\
        & $\alpha-$SPIN ($h=1$) & $49.40$ & $33.14$ & $65.82$ & $17.66$ & $72.25$ & $41.13$ & $46.57$ \\
        \midrule
        %\multirow{3}{*}{\shortstack[l]{Iteration 2}} & SPIN & $50.34$ & $33.91$ & $67.01$ & $20.24$ & $73.82$ & $40.65$ & $47.66$  \\
        \multirow{2}{*}{Iteration 2} & $\alpha-$SPIN ($h=2$) & $50.09$ & $33.86$ & $66.22$ & $21.23$ & $73.65$ & $40.54$ & $47.60$ \\
        & $\alpha-$SPIN ($h=1$) & $49.40$ & $33.31$ & $65.82$ & $18.12$ & $72.23$ & $40.93$ & $46.69$ \\
        \bottomrule
    \end{tabular}
    }
\end{table}

\begin{comment}
\begin{figure}[h!]
    \centering    \includegraphics[scale=0.5]{pic/SPIN_vs_AlphaSPIN_v4.png}
    \caption{MT bench of $\alpha$-SPIN ($h=2$) versus original SPIN. We compare SPIN and $\alpha$-SPIN across 3 iterations.}
\label{fig:SPINvsAlphaSPIN}
\end{figure}
\end{comment}

\begin{figure}[h!]
  \centering
  \begin{minipage}{.5\textwidth}
    \centering
    \includegraphics[width=1\linewidth]{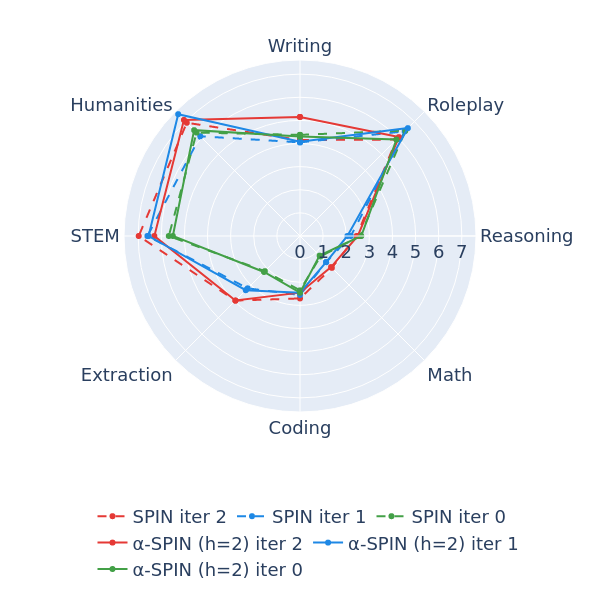}
    \caption{$\alpha$-SPIN ($h=2$) vs vanilla SPIN.}
\label{fig:SPINvsAlphaSPIN}
  \end{minipage}%
  \begin{minipage}
  {.5\textwidth}
    \centering
    \includegraphics[width=1\linewidth]{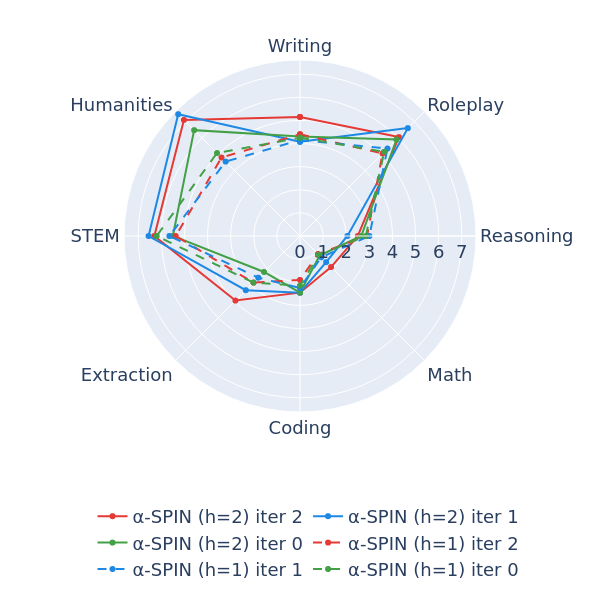}
    \caption{$\alpha$-SPIN ($h=2$) vs $\alpha$-SPIN ($h=1$).}
\label{fig:alphaSPINvsPrevious}
  \end{minipage}
\end{figure}

From the chart in Figure \ref{fig:SPINvsAlphaSPIN}, we notice that all iterations of SPIN and $\alpha$-SPIN models exhibit a high degree of consistency in Math, Coding, and Extraction, suggesting robust capabilities in structured problem-solving and information processing. Moreover, it is notable that the $\alpha$-SPIN models, represented by solid lines, seem to outperform the standard SPIN models, shown with dashed lines, particularly in the Humanities and STEM domains, which may indicate enhanced contextual understanding or domain-specific training. Furthermore, while the iterations $1$ and $2$ of SPIN show improvement over iteration $0$ in the Reasoning and Roleplay domains, the iteration $2$ of $\alpha$-SPIN demonstrates the most substantial gain, suggesting iteration-specific enhancements.

\subsection{Investigating the effect of the history length}

\begin{comment}

\begin{figure}[h!]
    \centering    \includegraphics[scale=0.5]{pic/AlphaSPIN_vs_AlphaSPIN(t-1)_v5.png}
    \caption{$\alpha$-SPIN with arithmetic generation vs $\alpha$-SPIN with generation from the previous model.}
\label{fig:alphaSPINvsPrevious}
\end{figure}

\end{comment}

The chart in Figure \ref{fig:alphaSPINvsPrevious} highlights the importance of going beyond one previous iterate in order to generate loser answers $y_l$ for the DPO loss, similar to what was done in the official SPIN implementation.

The chart in Figure \ref{fig:alphaSPINvsPrevious} suggests that iterations 0 and 1 of $\alpha$-SPIN with $h=2$ and $\alpha$-SPIN with $h=1$ perform similarly across most domains, with slight variations. Moreover, iteration 2 of both $\alpha$-SPIN ($h=2$) and $\alpha$-SPIN ($h=1$) shows a noticeable improvement in areas such as Writing, Humanities, and STEM.

Table \ref{tab:h12} confirms the benefit of the mixing trick ($h=2$) harnessed in the SPIN official implementation. Indeed, we observe that the empirical improvement of using $h=2$ over $h=1$ also transfers to our $\alpha$-SPIN approach.

\subsection{Investigating the effect of fictitious play}

\begin{figure}[h!]
  \centering
  \begin{minipage}{.5\textwidth}
    \centering
    \includegraphics[width=1\linewidth]{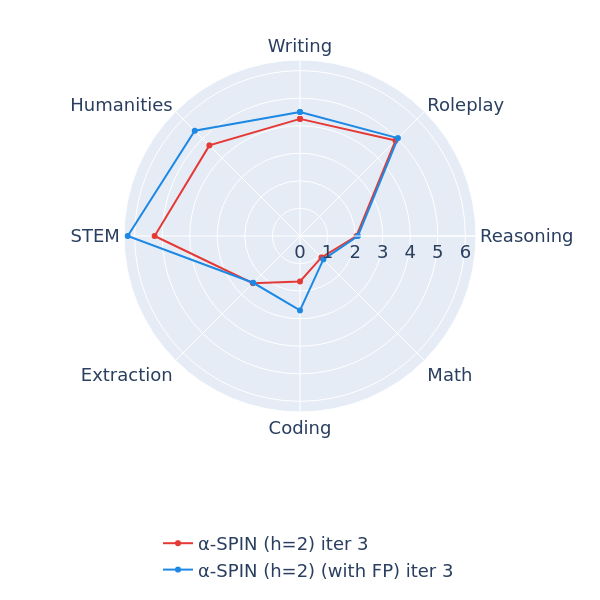}
    \caption{Fictitious play effect on $\alpha$-SPIN.}
    \label{fig:fig4}
  \end{minipage}%
  \begin{minipage}
  {.5\textwidth}
    \centering
    \includegraphics[width=1\linewidth]{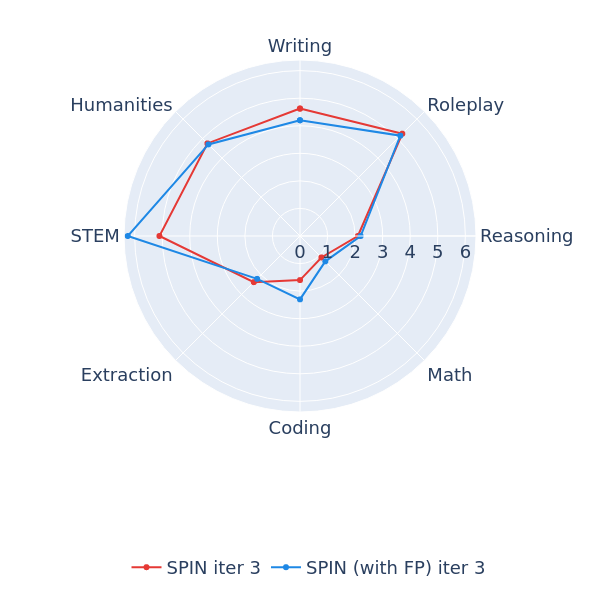}
    \caption{Fictitious play effect on SPIN.}
    \label{fig:fig5}
  \end{minipage}
\end{figure}

The paired radar charts in Figures \ref{fig:fig4} and \ref{fig:fig5} offer a visual comparison between $\alpha$-SPIN models with and without the incorporation of fictitious play, across a variety of cognitive domains. Figure \ref{fig:fig4} highlights the impact of fictitious play on $\alpha$-SPIN's performance, where it appears that the inclusion of this factor marginally enhances capabilities in Writing, Humanities, and STEM, while exhibiting a slight reduction in Reasoning. Figure \ref{fig:fig5} shows the influence of fictitious play on the standard SPIN models, where improvements are more pronounced in the domains of Writing, Roleplay, and STEM.

\begin{comment}

\begin{figure}[h]
    \centering    \includegraphics[scale=0.6]{pic/fictitiousAlphaPlaySPIN_v2.png}
    \caption{Fictitious Play effect on Alpha SPIN}
\label{fig:arithmeticVSoriginal}
\end{figure}

\begin{figure}[h]
    \centering    \includegraphics[scale=0.6]{pic/fictitiousPlaySPIN_v2.png}
    \caption{Fictitious Play effect on SPIN}
\label{fig:arithmeticVSoriginal}
\end{figure}

\end{comment}

\section{Conclusion}
\label{sec::conclusion}

In this paper, we investigated various ways of regularizing the self-play paradigm for the language model alignment problem.
We mainly explored two directions: (1) incorporating an additional KL-penalty to enforce the learned policies to remain close to the base model ; (2) smoothing the opponent policy across multiple previous iterates in order to keep track of the history of past strategies and avoid any abrupt deviation in the learning process.
We collected all these variations of SPIN into our $\alpha$-SPIN framework.

Our investigation revealed the following takeaways for improving the performance of SPIN via regularization: (i) the positive effect of the KL penalty term with respect to the base model, (ii) the improved performance induced by mixing several past policies in the sampler, and (iii) the promising results of the fictitious play approach.

Our initial results on using the geometric mixture as a sampler, obtained using GFlowNet-finetuning are indicative of the importance of GFlowNet hyperparameters. Future directions of research include replacing the geometric mixture that is used as reference policy in $\alpha$-SPIN with an arithmetic mixture, both as a sampler, which is trivial to sample from, and in the loss, even though that would be less interpretable in terms of KL regularization, or computing an exponential moving average (EMA) in the parameter space (i.e. on the $\theta$'s) as proposed in \citep{munos2023nash}.
Another promising line of research consists in designing a novel family of self-play alignment techniques by substituting the DPO loss with the IPO loss \citep{IPO} evoked in \ref{subsec:alternative}.
Indeed, the IPO approach comes with several advantages including the fact that it is less prone to overfitting than DPO, and it is linked with Nash-MD as shown in \citep{calandriello2024human}.

%\subsubsection*{Author Contributions}
%If you'd like to, you may include  a section for author contributions as is done
%in many journals. This is optional and at the discretion of the authors.

%\subsubsection*{Acknowledgments}
%Use unnumbered third level headings for the acknowledgments. All
%acknowledgments, including those to funding agencies, go at the end of the paper.

\bibliography{colm2024Conference}
\bibliographystyle{colm2024Conference}

\appendix

\newpage

\section{GFlowNet-fine-tuning to sample from the geometric mixture}
\label{sec::gflownets}
%\paragraph{Key observation}
We point out that the presence of the reference policy $\pi_{\text{ref}}$ in the DPO loss (Eq.\ref{eq:dpo_loss}) originates in the KL regularization term $\text{KL}(\pi(\cdot|x) \| \pi_{\text{ref}}(\cdot|x))$ in Eq.(\ref{eq:kl_reg}). This is \emph{consistent} with the fact that the DPO approach generates the answers $y_w,y_l$ from $\pi_{\text{ref}}$. From this perspective, the SPIN iterative method appears to be \emph{inconsistent} since the standard implementation generates the rejected answers $y_l$ from an arithmetic mixture $\frac12 ( \pi_{\theta_{t-1}}(\cdot|x) + \pi_{\theta_{t-2}}(\cdot|x) )$ between the two previous iterates (see Remark \ref{rmk:official_implem}), but minimizes a DPO loss function that only takes the previous policy $\pi_{\text{ref}} = \pi_{\theta_{t-1}}$ as reference. Consequently, in order to design a ``consistent'' version of $\alpha$-SPIN, we would need to sample the negative answers from $\pi_{\text{ref}}(\cdot|x) \propto \pi_{\theta_{t-1}}(\cdot|x)^\alpha \odot \pi_{\text{base}}(\cdot|x)^{1-\alpha}$,
i.e., a geometric mixture between the previous iterate and the base policy.

\paragraph{Sampling from the geometric mixture} 
% Similar to  the original version of SPIN, where the policy used in the DPO loss \ref{eq:dpo_loss} is the one used to generated the less prefered outputs ($y_l$), we investigated using the geometric mixture $\pi_{ref} \propto \pi_{\theta_{t-1}}^\alpha \odot \pi_{\text{base}}^{1-\alpha}$ as a sampling policy. 
Sampling from $\pi_{\text{ref}}(\cdot|x) \propto \pi_{\theta_{t-1}}(\cdot|x)^\alpha \odot \pi_{\text{base}}(\cdot|x)^{1-\alpha}$ poses an important challenge. Typically, generating a sequence from a language model is done auto-regressively: sampling $y = (y^1, \dots, y^L) \sim \pi(y \mid x)$ amounts to sampling $y^1 \sim \pi(y^1 \mid x)$, and subsequently $y^k \sim \pi(x, y^{<k})$ until some terminating token is sampled. This is because the joint decomposes as the product of the conditionals:
\begin{equation}
    \pi(y \mid x) = \pi(y^1 \mid x) \pi(y^2 \mid x, y^1) \dots \pi(y^L \mid x, y^{<L}).
\end{equation}
However, the geometric mixture of two auto-regressive sequence generators does \emph{not} decompose at the level of the tokens.
Indeed, given a sequence $y=(y^1,\dots,y^L)$ of length $L$, first notice that we have:
\begin{multline}
    \pi_{\theta_{t-1}}^\alpha(y|x) \times \pi_{\text{base}}^{1-\alpha}(y|x)
    = \left[ \prod_{1\le k\le L} \pi_{\theta_{t-1}}(y^k|x,y^{< k}) \right]^\alpha \times \left[ \prod_{1\le k\le L} \pi_{\text{base}}(y^k|x,y^{< k}) \right]^{1-\alpha} \\
    = \prod_{1\le k\le L} \pi_{\theta_{t-1}}(y^k|x,y^{< k})^\alpha \pi_{\text{base}}(y^k|x,y^{< k})^{1-\alpha} ,
\end{multline}
where $y^{<k}=(y^1,\dots,y^{k-1})$ denotes the $k$-th partial sequence (with convention $y^{<1}=\emptyset$).
However, the corresponding normalization constant is in general not equal to the product of the auto-regressive normalization constants for any given sequence $y$:
\begin{multline}
    \sum_{y'=(y^{'1},\dots,y^{'L})} \pi_{\theta_{t-1}}^\alpha(y'|x) \times \pi_{\text{base}}^{1-\alpha}(y'|x)
    = \prod_{k=1}^L \sum_{y^{'k}} \pi_{\theta_{t-1}}(y^{'k}|x,y^{'< k})^\alpha \pi_{\text{base}}(y^{'k}|x,y^{'< k})^{1-\alpha} \\
    \neq \prod_{k=1}^L \sum_{y^{'k}} \pi_{\theta_{t-1}}(y^{'k}|x,y^{< k})^\alpha \pi_{\text{base}}(y^{'k}|x,y^{< k})^{1-\alpha} . 
\end{multline}

The fact that the probability distribution $\frac{1}{Z}\pi_{\theta_{t-1}}^\alpha(y|x) \cdot \pi_{\text{base}}^{1-\alpha}(y|x)$ does not decompose as the product of token-wise geometric mixture probability distributions is actually why beam search and its variations \citep{vijayakumar2016diverse,shao2017generating} are used to generate high-likelihood sequence continuations, rather than auto-regressively sampling from tempered distributions, i.e., proportionally to $\pi(y^k \mid x, y^{<k})^\frac{1}{T}$.

Ideally, we would like our target distribution to factorize as a product of conditionals from which sampling is tractable. Essentially, this requires looking for conditional probability distributions $q(y^k \mid x, y^{<k})$ such that:
\begin{align}
    \prod_{1 \leq k \leq L} q(y^k \mid x, y^{<k}) \propto \pi_{\theta_{t-1}}^\alpha(y|x) \times \pi_{\text{base}}^{1-\alpha}(y|x).
\end{align}

Generative Flow Networks \citep[GFlowNets;][]{bengio2023gflownet} have been introduced to tackle this very problem. In fact, \cite{hu2023amortizing} used GFlowNets to amortize sampling from intractable distributions, including those of the type $\propto \pi(y^k \mid x, y^{<k})^\frac{1}{T}$. Similarly, we use the proposed GFlowNet-finetuning algorithm to turn $\pi_{\theta_{t-1}}$ into a sampler of sequences with likelihoods proportional to the geometric mixture.

\paragraph{More details on GFlowNet-finetuning for the geometric mixture}
GFlowNet-finetuning requires the specification of a non-negative reward function $y \mapsto R(y \mid x)$ over the space of sequences $\gY$. This is actually a special case of conditional GFlowNets \citep{bengio2023gflownet}, where the conditioning variable is the prompt $x$. Essentially, training a conditional GFlowNet leads to an amortized conditional sampler from the family of distributions $\left(y \mapsto \frac{1}{Z_x} R(y \mid x)\right)_{x}$, where $Z_x$ is the normalizing constant.

Therefore, the reward function needed to obtain a sampler from $\pi_{ref}$ is $R(y \mid x) = \pi_{\theta_{t-1}}^\alpha(y|x) \times \pi_{\text{base}}^{1-\alpha}(y|x).$

The state space is the set of partial sequences, and transitions are only defined between two states that differ by one token. States that end with a termination token have no children in the corresponding directed acyclic graph, and correspond to the sample space $\gY$ (the terminating states in GFlowNet parlance).

Similar to \cite{hu2023amortizing}, we use the modified version of the sub-trajectory balance loss \citep{madan2023learning} that accounts for trajectories being terminable at all states (as long as the termination token is appended).

To train a conditional GFlowNet, choosing a set of conditioning variables from which learning trajectories are generated is important. In our setting, this amounts to choosing the right dataset of prompts. We investigated using (1) the same dataset used in the sentence continuation task of \cite{hu2023amortizing}, i.e. prompts from the OpenWebText corpus \citep{Gokaslan2019OpenWeb}, but also (2) the training set of the SFT dataset used in our algorithm, i.e. prompts from the UltraChat 200k dataset \citep{Ultrachat}. 

Similar to \cite{hu2023amortizing}, we used 1000 prompts randomly sampled from these datasets, and used LoRA \cite{hu2021lora} with rank 16 to finetune $\pi_{\theta_{t-1}}$, at each iteration, we finetuned for 10 epochs, with a learning rate of $10^{-4}$.

\begin{figure}[h!]
    \centering    \includegraphics[scale=0.6]{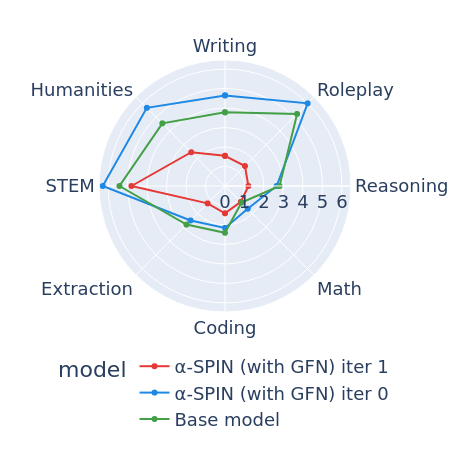}
    \caption{Results on GFlowNets}
\label{fig:arithmeticVSoriginal2}
\end{figure}

\paragraph{Experimental results}
The chart in Figure \ref{fig:arithmeticVSoriginal2} shows that the performance degrades starting from Iteration 1, and that more effort should be spent on the specifics of GFlowNet-finetuning in order to get accurate approximate samplers of the geometric mixture.

\end{document}